# A Monte-Carlo Algorithm for Dempster-Shafer Belief

Nic Wilson
Department of Computer Science
Queen Mary and Westfield College
Mile End Rd., London E1 4NS, UK

## Abstract

A very computationally-efficient Monte-Carlo algorithm for the calculation of Dempster-Shafer belief is described. If Bel is the combination using Dempster's Rule of belief functions $\text{Bel}_1, \ldots, \text{Bel}_m$ then, for subset $b$ of the frame $\Theta$, $\text{Bel}(b)$ can be calculated in time linear in $|\Theta|$ and $m$ (given that the weight of conflict is bounded). The algorithm can also be used to improve the complexity of the Shenoy-Shafer algorithms on Markov trees, and be generalised to calculate Dempster-Shafer Belief over other logics.

## 1 INTRODUCTION

One of the major perceived problems with application of the Dempster-Shafer Theory [Shafer, 76] has been its apparent computational complexity e.g., [Kyburg, 87], [Bonissone, 87]. This is because the Dempster-Shafer theory as usually implemented involves repeated application of Dempster's Rule of Combination, keeping a record at each stage of each subset of $\Theta$ with a non-zero mass. For example the combination of $m$ simple support functions can have as many as $2^q$ non-zero masses where $q$ is the minimum of $m$ and $|\Theta|$, thus making the approach computationally infeasible for large $m$ and $|\Theta|$.

There have been a number of schemes to deal with this; [Barnett, 81] showed how calculation of Dempster-Shafer belief in a very special case, when all the evidence sets are either singletons or complements of singletons, belief could be calculated in linear time. [Gordon and Shortliffe, 85] extended this with an efficient approximation to Dempster-Shafer belief for hierarchically related evidences, and it was shown in both [Shafer and Logan, 87] and [Wilson, 87] that the hierarchical case could be dealt with exactly in a computationally efficient manner. The Shafer-Logan algorithm was generalised to propagation of belief functions in Markov Trees [Shafer and Shenoy, 88] but, although this is a very important contribution, it still requires that the product space associated with the largest clique is small, a condition which will by no means always be satisfied. The hierarchical evidence algorithm in [Wilson, 87] was generalised to arbitrary evidence sets [Wilson, 89] and, because it calculates belief directly without first calculating the masses, it leads to very substantial increases in efficiency (see section 4). However this algorithm appears to have complexity worse than polynomial, which is not surprising since Dempster's Rule is $\#P$-complete [Orponen, 90], [Provan, 90].

This paper describes the Monte-Carlo algorithm given in [Wilson, 89] which also calculates belief directly (or, more accurately, it approximates belief up to arbitrary accuracy). This calculation has very low complexity, showing that the general pessimism about the complexity of Dempster-Shafer Theory is misguided. The use of Monte-Carlo algorithms for calculating Dempster-Shafer belief has also been suggested in [Pearl, 88], [Kämpke, 88] and [Kreinovich and Borrett, 90].

## 2 THE MONTE-CARLO ALGORITHM

Let $\text{Bel}_1, \ldots, \text{Bel}_m$ be belief functions on a finite frame $\Theta$, and let $\text{Bel} = \text{Bel}_1 \oplus \cdots \oplus \text{Bel}_m$ be their combination using Dempster's Rule. Using the model of [Dempster, 67] $\text{Bel}_i$ is represented by a probability function $P_i$ (on a finite set $\Omega_i$) and a compatibility function $\Gamma_i : \Omega_i \mapsto 2^\Theta$ where the meaning of $\Gamma_i$ is 'for $\tau \in \Omega_i$, if $\tau$ is true then so is $\Gamma_i(\tau)$'.

The mass function $m_i$ is given by: for $\varepsilon_i \in \Omega_i$, $m_i(\Gamma_i(\varepsilon_i)) = P_i(\varepsilon_i)$, and, for $b \subseteq \Theta$,

$$\text{Bel}_i(b) = P_i(\Gamma_i(\varepsilon_i) \subseteq b), \text{ that is, } \sum_{\varepsilon_i : \Gamma_i(\varepsilon_i) \subseteq b} P_i(\varepsilon_i).$$

Let $\Omega = \Omega_1 \times \cdots \times \Omega_m$ and for $\varepsilon = (\varepsilon_1, \ldots, \varepsilon_m)$ define $\Gamma(\varepsilon) = \bigcap_{i=1}^m \Gamma_i(\varepsilon_i)$. Define the 'independent probability function' P' on $\Omega$ by $P'((\varepsilon_1, \ldots, \varepsilon_m)) = \prod_{i=1}^{m} P_i(\varepsilon_i)$.

Using [Dempster, 67] it can be seen that

$$\text{Bel}(b) = P'(\Gamma(\varepsilon) \subseteq b | \Gamma(\varepsilon) \neq \emptyset),$$

where e.g. $P'(\Gamma(\varepsilon) \neq \emptyset)$ just means $\sum_{\varepsilon : \Gamma(\varepsilon) \neq \emptyset} P'(\varepsilon)$. $\Gamma(\varepsilon)$ can be viewed as a random set [Nguyen, 78].

The Monte-Carlo algorithm just simulates the last equation.

A large number, $N$, of trials are performed. For each trial:

1. Randomly pick $\varepsilon$ such that $\Gamma(\varepsilon) \neq \emptyset$:
   a. For $i = 1, \ldots, m$
      randomly pick an element of $\Omega_i$, i.e.
      pick $\varepsilon_i$ with probability $P_i(\varepsilon_i)$
      Let $\varepsilon = (\varepsilon_1, \ldots, \varepsilon_m)$
   b. If $\Gamma(\varepsilon) = \emptyset$ then restart trial;
2. If $\Gamma(\varepsilon) \subseteq b$ then trial succeeds, let $T = 1$
   else trial fails, let $T = 0$

The proportion of trials that succeed converges to Bel($b$):

$E[T] = P'(\Gamma(\varepsilon) \subseteq b | \Gamma(\varepsilon) \neq \emptyset) = \text{Bel}(b)$.

$\text{Var}[T] = E[T^2] - (E[T])^2 = E[T] - (E[T])^2 = \text{Bel}(b)(1 - \text{Bel}(b)) \leq \frac{1}{4}$.

Let $\bar{T}$ be the average value of $T$ over the $N$ trials, i.e., the proportion of trials that succeed.

$$E[\bar{T}] = \frac{N E[T]}{N} = \text{Bel}(b)$$
$$\text{and} \quad \text{Var}[\bar{T}] = \frac{N \text{Var}[T]}{N^2} \leq \frac{1}{4N}$$

Therefore the variance (and so also the standard deviation) for the estimate, $\bar{T}$, of Bel($b$) can be made arbitrarily small independently of $|\Theta|$ and $m$.

Let us say that the estimate $\bar{T}$ of Bel($b$) 'has accuracy $k$' if 3 standard deviations of $\bar{T}$ is less than or equal to $k$. Then $\bar{T}$ has accuracy $k$ if $N \geq \frac{9}{4k^2}$.

Testing separately if $\Gamma(\varepsilon) = \emptyset$ and if $\Gamma(\varepsilon) \subseteq b$ wastes time; these tests can be combined within the same algorithm (where $x_j$ denotes the $j$th element of $\Theta$):

For each trial:

**repeat**
  pick $\varepsilon$ with probability $P'(\varepsilon)$
  $T_\emptyset := 1; T := 1$
  **for** $j = 1$ to $|\Theta|$
    **if** $\Gamma(\varepsilon) \ni x_j$ **then**
      $T_\emptyset := 0;$   (since $\Gamma(\varepsilon) \neq \emptyset$)
      **if** $x_j \notin b$ **then** $T := 0$; exit trial;
    **end if**   (since $\Gamma(\varepsilon) \not\subseteq b$)
  **end if**
  next $j$
**until** $T_\emptyset = 0$

## 3 COMPUTATION TIME

Picking $\varepsilon$ involves $m$ random numbers so takes less than $Am$ where $A$ is constant, approximately the time it takes to generate a random number (with efficient storing of the $P_i$s). Testing if $\Gamma(\varepsilon) \ni x_j$ takes less than $Bm$ for constant $B$. For a given trial there is a probability $\kappa = P'(\Gamma(\varepsilon) = \emptyset)$ that the **repeat-until** loop will be entered a second time. The expected number of **repeat-until** loops per trial is $\frac{1}{1-\kappa}$; $\kappa$ is a measure of the conflict of the evidences [Shafer, 76, p65].

Thus the expected time the algorithm takes is less than $\frac{N}{1-\kappa} m(A + B|\Theta|)$, and so the expected time to achieve accuracy $k$ is less than $\frac{9}{4(1-\kappa)k^2} m(A + B|\Theta|)$.

At least for the case where the Bel$_i$s are simple support functions, the condition $\Gamma(\varepsilon) \ni x_j$ can be tested more efficiently; under weak conditions this leads to expected time of less than $\frac{9}{4(1-\kappa)k^2}(Am + C|\Theta|)$ for constant $C$ [Wilson, 89].

## 4 EXPERIMENTAL RESULTS

The algorithm for the case where the Bel$_i$s are simple support functions has been implemented and tested using the language Modula-2 on a SUN 3/60 workstation. The results showed that the value of $A$ is much bigger than the value of $C$ in this implementation, $A$ being roughly $\frac{1}{4000}$ seconds and $C$ roughly $\frac{1}{50,000}$ seconds. $A$ is essentially the time taken to generate a random number, and $\frac{1}{4000}$ seconds seems rather slow for that. This suggests that very substantial speedups (of perhaps an order of magnitude or two) could be achieved by careful choice and use of the random number generator and the use of antithetic runs (so that the random number generator is only used once for several different data items).

The results indicate that, unless the evidences are extremely conflicting, the Monte-Carlo algorithm is practical for problems with large $m$ and $\Theta$. For example, with $\kappa = 0.5$, $m = |\Theta| = 40$, and with 1000 trials, the calculation of the approximate value of Bel($b$) would be expected to take 20.6 seconds. The 1000 trials mean that the standard deviation is less than 0.016, and so the confidence interval for the correct value of belief corresponding to 3 standard deviations would be roughly $[b - 0.05, b + 0.05]$. If instead we did 10,000 trials this would take a little over 3 minutes, and give a standard deviation of 0.005. Extrapolating the figures (which seems unlikely to cause problems in this case) gives an approximate time of 1 minute for $m = |\Theta| = 120$, with 1000 trials, and 5 minutes for $m = |\Theta| = 600$.

Also in [Wilson, 89] an exact algorithm for calculating belief is described (related to those described in [Provan, 90]) which involves expressing the event $\Gamma(\varepsilon) \subseteq b$ as a boolean expression and then calculating the probability of this using the laws of boolean algebra. Again this avoids explicit calculation of the masses. The complexity for the simple support function case appears to be approximately of the form $|\Theta|^{\log m}$.

The usual approaches for calculating belief are mass-based: they calculate the combined mass function and use this to calculate the appropriate belief (a good one of these is the fast Möbius transform in [Kennes and Smets, 90]). For large $m$ and $\Theta$ this is of necessity very computationally expensive, since if $q = \min(m, |\Theta|)$, there can be as many as $2^q$ masses. For simplicity it is assumed that the calculation of belief then just does $2^q$ REAL multiplications. The speed of REAL multiplication was tested on the same workstation and within the same language that the Exact and Monte-Carlo algorithms were tested on and implemented on and it was found that it did just over $10^4$ REAL multiplications per second. This gives the following results:

| $m, n$ | MC | Exact | Mass-based $\geq$ |
|---|---|---|---|
| 15 × 15 | 7 secs | 9 secs | 3 secs |
| 20 × 20 | 11 secs | 13 secs | 1 min |
| 25 × 25 | 13 secs | 46 secs | 1 hour |
| 30 × 30 | 15 secs | 3 mins | 1 day |
| 35 × 35 | 17 secs | 8 mins | 1 month |
| 50 × 50 | 25 secs | 2 hours | 3000 years |

The values for the Monte-Carlo algorithm were based on doing 1000 trials and the contradiction being 0.5. The figure of 2 hours for the Exact in the 50 case is a very rough upper bound derived from insufficient data.

Details of the experiments and the full results and analysis are given in [Wilson, 90b].

## 5 THE GENERALISED ALGORITHM

The algorithm can be generalised to deal with arbitrary logics [Wilson, 90a]. Let $L$ be the language of some logic. For each $i$, $\text{Bel}_i$ is now a function from $L$ to $[0, 1]$ saying how much the evidence warrants belief in propositions in $L$ and the compatibility function is a function $\Gamma_i : \Omega_i \mapsto L$. The combined compatibility function $\Gamma$ is now defined by

$$\Gamma((\varepsilon_1, \ldots, \varepsilon_m)) = \bigwedge_{i=1}^{m} \Gamma_i(\varepsilon_i),$$

(or $\Gamma((\varepsilon_1, \ldots, \varepsilon_m)) =$ the set $\{\Gamma_i(\varepsilon_i) : i = 1, \ldots, m\}$ if the logic doesn't have conjunction).

For each trial:

1. Randomly pick $\varepsilon$ such that
    $\Gamma(\varepsilon)$ is not contradictory:
    a. For $i = 1, \ldots, m$
        randomly pick an element of $\Omega_i$, i.e.
        pick $\varepsilon_i$ with probability $P_i(\varepsilon_i)$
    Let $\varepsilon = (\varepsilon_1, \ldots, \varepsilon_m)$
    b. If $\Gamma(\varepsilon)$ is contradictory then restart trial;
2. If $b$ can be deduced from $\Gamma(\varepsilon)$
    then trial succeeds, let $T = 1$
    else trial fails, let $T = 0$

Undecidability and semi-decidability would clearly cause problems, in which case trials which went on for too long would have to be cut short; if $T$ for these trials was given the value 0 then this would lead to a lower bound for Bel($b$). This technique of prematurely halting trials that take too long could be used to increase the efficiency for other cases as well, at the cost of only finding lower and upper bounds for Bel($b$).

The time this algorithm takes is then approximately $\frac{N}{1-\kappa}(Am + R)$ where $R$ is the average time it takes to see if $\Gamma(\varepsilon)$ is contradictory, and if $\Gamma(\varepsilon)$ allows $b$ to be deduced. Given that the weight of conflict of the evidences is bounded this means that the complexity is proportional to that of proof in the logic; it is hard to see how any sensible uncertainty calculus could do better than this (although the complexity for this Monte-Carlo algorithm has a very large constant term if high accuracy is required).

As Shafer points out [Shafer, 90] $|\Theta|$ can be a large product space, making the first algorithm impractical. The generalised algorithm can also be used to greatly improve the complexity of the algorithms for calculating Belief in Markov trees [Shafer and Shenoy, 88]. For each trial, propositions (i.e. belief functions with a single focal element) must be propagated through the Markov tree. The complexity is then proportional to that of propagating propositions, rather than the whole belief functions. Some other propositional cases have been dealt with in [Wilson, 89].

## 6 DISCUSSION

There are two obvious drawbacks with the Monte-Carlo algorithm:

(i) if very high accuracy is required then the Monte-Carlo algorithm will require a large number of trials (quadratic in the reciprocal of accuracy) so giving a very high constant factor to the complexity;

(ii) when the evidence is highly conflicting the Monte-Carlo algorithm loses some of its efficiency. I don't see

this as a great problem since an extremely high weight of conflict would suggest, except in exceptional circumstances, that Dempster's Rule is being applied when it is not valid, e.g. updating a Bayesian prior with a Dempster-Shafer belief function [see Wilson, 91]. I also argue there that, although Dempster's Rule has strong justifications for the combination of a finite number of simple support functions, the more general case has not been convincingly justified: the Monte-Carlo algorithm is guaranteed to give results in accordance with Dempster's Rule, but it remains to be seen if these are always sensible.

It may be important to know which relatively small sets have relatively high beliefs: the Monte-Carlo algorithm can be easily applied to deal with this problem.

Dempster's Rule makes particular independence assumptions, using a single probability function on $\Omega$. By modifying step 1 of the algorithms the beliefs corresponding to other probability functions on $\Omega$ can be calculated.

### Acknowledgements

I am currently supported by the ESPRIT basic research action DRUMS (3085). Most of the material in this paper was produced in the period Summer '87-Summer '88 when I was employed by the The Hotel and Catering Management, and Computing and Mathematical Sciences Departments of Oxford Polytechnic. Thanks also to Bills Triggs and Boatman for their help during this period, and more recently to Mike Clarke.